\documentclass[11pt]{article}

\usepackage[preprint]{acl}

\usepackage{times}
\usepackage{latexsym}
\usepackage{booktabs}
\usepackage{amsmath}
\usepackage{amssymb}
\usepackage{multirow}
\usepackage{xcolor}
\usepackage{url}
\usepackage{array} 
\usepackage{algorithm}
\usepackage{algorithmic}
\usepackage{colortbl}

\usepackage[T1]{fontenc}
\usepackage[utf8]{inputenc}

\usepackage{microtype}

\usepackage{inconsolata}

\usepackage{graphicx}

\usepackage{booktabs,amsmath,amssymb,array,graphicx,xcolor}
\newcommand{\method}{Jag}
\newcommand{\acc}{Acc@0.5}
\title{Jag: Direct Box Prediction for Efficient Visual Grounding}

\author{
 \textbf{Xiuyuan Zhu\textsuperscript{1,2}},
 \textbf{Ke Lu\textsuperscript{1,3}},
 \textbf{Hao Wu\textsuperscript{1,2}},
\\
 \textbf{Zijin Du\textsuperscript{1}},
 \textbf{Dongming Zhang\textsuperscript{2}},
 \textbf{Jian Xue\textsuperscript{1,*}}
\\
\\
 \textsuperscript{1}University of Chinese Academy of Sciences, Beijing, China
\\
 \textsuperscript{2}State Key Laboratory of Communication Content Cognition, Beijing, China
\\
 \textsuperscript{3}Peng Cheng Laboratory, Shenzhen, Guangdong, China
\\
 \small{
   \textsuperscript{*}Correspondence: xuejian@ucas.ac.cn
 }
}

\begin{document}
\maketitle
\begin{abstract}
Visual grounding combines image--language understanding with geometric
prediction, yet autoregressive multimodal models often express a box
as a coordinate sequence. Sequential decoding adds computation to the
multimodal processing needed for localization.
Adapting these representations to box geometry offers a route to
direct prediction.
We present \method{}, which adapts a compact pretrained multimodal
model for direct continuous box prediction. A lightweight regression
head reads the last valid input-token state, conditioned on the image
and referring expression, and predicts all four box coordinates together.
Geometric supervision jointly adapts the language backbone, visual
merger, and regression head, allowing localization losses to shape both
the multimodal representations and the coordinate mapping.
Inference retains image--language processing and returns the box in
one forward pass, without coordinate generation or text parsing.
Across benchmarks, \method{}
consistently improves grounding accuracy over its base model across
the observed target sizes and expression groups. Component ablations
show that jointly adapting the representations and head substantially
improves accuracy over training the head alone.
Against existing specialized grounding models, it achieves
competitive accuracy with lower single-request latency and GPU memory
use. These results show how supervision tailored to the geometric
answer can specialize a compact pretrained multimodal model for
accurate grounding and efficient direct prediction. 
\href{https://xyzzzh.github.io/Jag/}{\textcolor{blue}{Project page}}.
\end{abstract}

\section{Introduction}
\label{sec:introduction}

Visual grounding couples image--language interpretation with a compact
geometric answer. For an expression such as ``elephant with trunk up,''
the raised trunk identifies the referent, while the required box
delimits the animal's extent.
Multimodal models that use autoregressive coordinate generation produce
text that is parsed into a bounding box. Sequential decoding and parsing
add output computation to the image--text processing needed for localization.

Continuous box prediction is established in TransVG and MDETR
\citep{deng2021transvg,kamath2021mdetr}. NExT-Chat extends it to a
multimodal language model by regressing from a location-token state
\citep{zhang2024nextchat}, while LocateAnything learns parallel
coordinate prediction within box-aligned token blocks
\citep{wang2026locateanything}. These models offer alternatives
to coordinate-by-coordinate generation. We build on this work to
adapt a compact pretrained model for
direct box prediction from an existing input-token state.

Motivated by task-specific prediction and continuous box regression
\citep{almeida2026jev,deng2021transvg,kamath2021mdetr,zhang2024nextchat},
we aim to retain pretrained image--language processing while predicting
the box directly in a single forward pass.

Adapting such a model involves connecting its image--language
representations to geometric supervision. The readout must reflect
both the image and the referring expression so that the predicted box
is conditioned on the requested target.
Pretrained hidden states are learned for language generation, and
fitting a regression head to fixed features confines adaptation to the
coordinate mapping. Joint adaptation allows box supervision to shape
both the multimodal representations and their mapping to coordinates.

We present \method{}, which adapts Qwen3.5-0.8B
\citep{qwen3.5} for direct bounding-box prediction.
A lightweight multilayer perceptron reads the last valid input-token
state and predicts continuous box centers and dimensions.
This position follows the full visual and textual input, so box
prediction reuses the same multimodal computation.
L1 and GIoU losses supervise coordinate values and geometric overlap.
These losses guide the joint adaptation of the language backbone,
visual merger, and regression head for localization.
The model returns the complete box in one multimodal forward pass,
without subsequent coordinate generation or text parsing.
Our qualitative analysis revisits the elephant example to examine the
resulting object localization.

Our contributions are summarized as follows:
\begin{enumerate}
\item We introduce \method{}, which adapts a compact pretrained
multimodal model for continuous box prediction from an existing
input-token state. It returns the complete bounding box in a single
forward pass.
\item We combine geometric supervision with joint adaptation of the
language backbone, visual merger, and regression head. Component
ablations show that joint adaptation substantially improves accuracy
over training the regression head alone.
\item Across the evaluated benchmarks, \method{} improves grounding accuracy
over Base and NExT-Chat and is competitive with LocateAnything.
It achieves lower single-request latency and GPU memory use than
the compared specialized grounding models.
\end{enumerate}

\section{Related Work}
\label{sec:related}

\paragraph{Visual grounding and continuous regression.}
MAttNet uses modular language attention to match expressions with
candidate regions \citep{yu2018mattnet}. One-stage approaches fuse
language with dense visual features to localize referents
\citep{yang2019faoa,yang2020resc,huang2021lbyl}.
DETR predicts boxes in parallel from learned object queries;
MDETR and Grounding DINO extend detector architectures with language
conditioning \citep{carion2020detr,kamath2021mdetr,liu2025groundingdino}.
TransVG and TransVG++ use transformer representations and a dedicated
regression token for direct box prediction
\citep{deng2021transvg,deng2023transvgpp}.
Within multimodal language models, NExT-Chat maps a generated trigger's
hidden state through an MLP to a continuous box, using L1 and GIoU
supervision \citep{zhang2024nextchat}. \method{} retains this regression
approach but reads the last valid input-token state, returning the box
without generating a trigger (Table~\ref{tab:related_methods}).

\begin{table}[!htbp]
\centering\small

\begingroup%
\setlength{\tabcolsep}{6pt}%
\renewcommand{\arraystretch}{1.16}%
\begin{tabular*}{\linewidth}{@{\extracolsep{\fill}}lll@{}}
\toprule
\textbf{Method} & \textbf{State source} & \textbf{Box output} \\
\midrule
NExT-Chat & Generated trigger & Continuous \\
LocateAnything & Box-aligned blocks & Discrete \\
\method{} & Existing input state & Continuous \\
\bottomrule
\end{tabular*}%
\endgroup%

\caption{Box prediction in closely related models
\citep{zhang2024nextchat,wang2026locateanything}.
NExT-Chat and \method{} use MLP heads for continuous box regression.
LocateAnything predicts discrete coordinates in parallel within each box block.}
\label{tab:related_methods}
\end{table}

\paragraph{Grounding in multimodal models.}
OFA incorporates grounding into a shared sequence-to-sequence framework
\citep{wang2022ofa}. Visual instruction tuning connects image
understanding with language generation in LLaVA and LLaVA-1.5
\citep{liu2023llava,liu2024llava15}, while InternVL3 and DeepSeek-VL2
advance multimodal understanding and conversation
\citep{zhu2025internvl3,wu2024deepseekvl2}.
For grounding, Shikra and Qwen-VL represent locations as coordinate
strings within language outputs \citep{chen2023shikra,bai2023qwenvl};
Qwen2.5-VL and Qwen3-VL likewise generate bounding-box or point
coordinates as text responses \citep{bai2025qwen25vl,bai2025qwen3vl}.
Kosmos-2 links text spans to discrete location tokens
\citep{peng2024kosmos2}. Ferret combines coordinate information with
continuous region features for referring inputs and generates box
coordinates in its responses \citep{you2024ferret}.
Recent work strengthens localization through task-specific
post-training, including Visual-RFT, VLM-R1, and Rex-Omni
\citep{liu2025visualrft,shen2025vlmr1,jiang2025rexomni}.
DeepGrounder extends this direction to remote-sensing visual grounding
through supervised and reinforcement fine-tuning
\citep{zhang2026deepgrounder}.
\method{} adapts multimodal representations for direct continuous
single-box prediction from an existing input-token state.

\paragraph{Coordinate representations and parallel decoding.}
Pix2Seq and SeqTR cast detection and grounding as autoregressive
prediction of quantized coordinate sequences
\citep{chen2022pix2seq,zhu2022seqtr}. Rex-Omni uses dedicated coordinate
tokens \citep{jiang2025rexomni}, while Hi-Token decomposes coordinates
into axis-specific digit tokens \citep{zhu2026hitoken}.
LocateAnything predicts coordinates in parallel within box-aligned
blocks, retains causal dependencies between blocks, and jointly learns
next-token and multi-token objectives \citep{wang2026locateanything}.
\method{} predicts four continuous values together from a shared input
state. Our parallel categorical control uses independent coordinate
heads; LocateAnything models within-block interactions through attention.

\paragraph{Geometric supervision and adaptation.}
GIoU provides an overlap-based loss for continuous box regression
\citep{rezatofighi2019giou}. For generated coordinate outputs,
Visual-RFT and VLM-R1 use IoU-based rewards in reinforcement fine-tuning
\citep{liu2025visualrft,shen2025vlmr1}.
IoU-PD combines coordinate-string supervision with geometry- and
reliability-weighted distillation from a privileged training-time teacher
\citep{zhu2026ioupd}. Hi-Token's Hi-GAR stage applies geometry-aware
GRPO after supervised fine-tuning \citep{zhu2026hitoken}.
\method{} directly supervises continuous predictions with L1 and GIoU
losses while adapting the language backbone, visual merger, and regression
head. The component and output-formulation ablations assess the effects
of representation adaptation and box formulation.

\section{Method}
\label{sec:method}

\begin{figure*}[t]
\centering
\includegraphics[width=\textwidth]{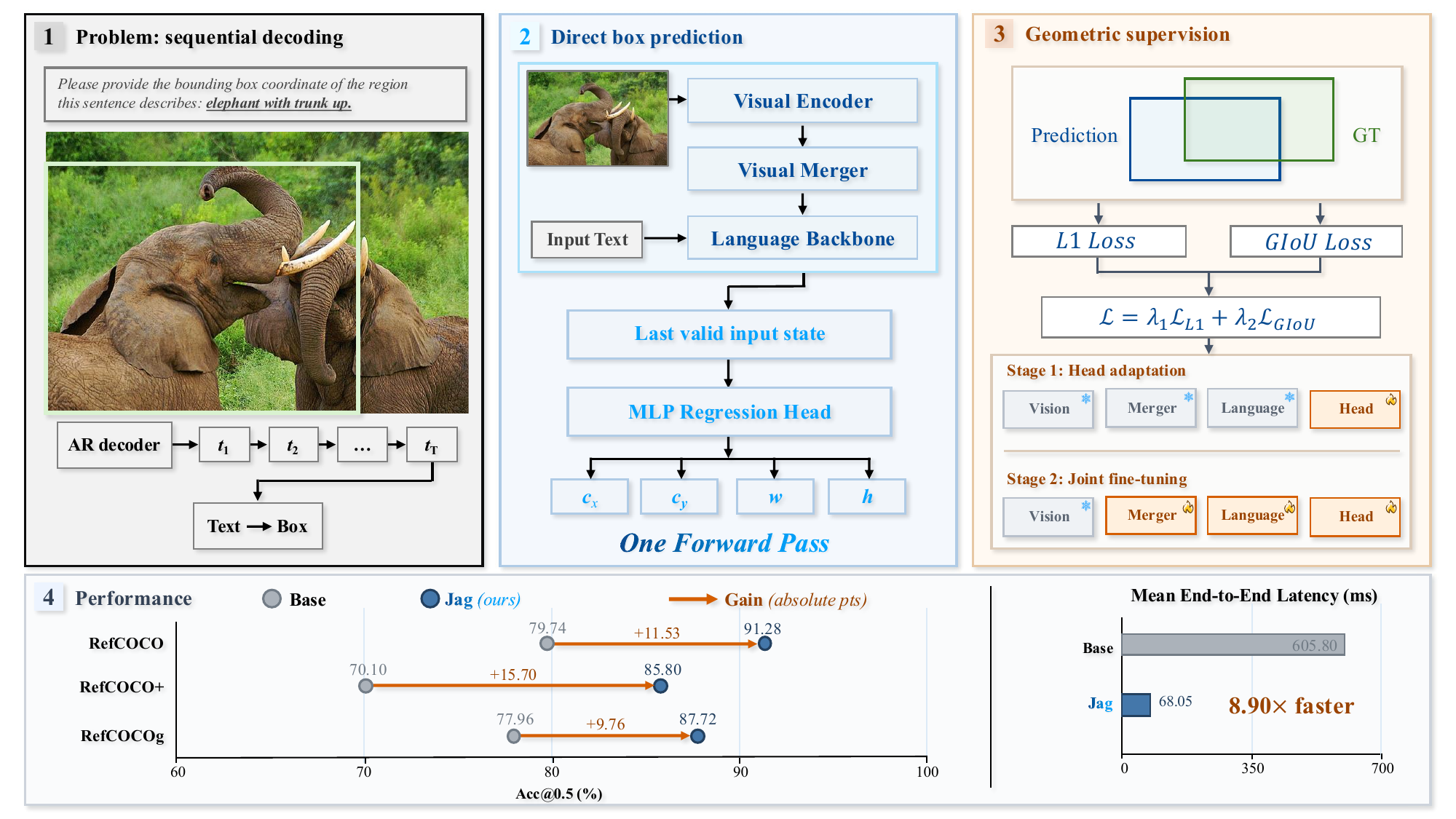}
\caption{\method{} adapts a compact multimodal model for accurate and
efficient visual grounding. It retains multimodal input
processing and replaces sequential coordinate generation with continuous
box prediction in one forward pass. A lightweight regression head reads
the last valid input state; geometric supervision jointly adapts the
head, language backbone, and visual merger for localization. The resulting
task adaptation improves grounding accuracy over the base model,
while the direct output reduces inference latency relative to
autoregressive localization.}
\label{fig:framework}
\end{figure*}

\method{} adapts Qwen3.5-0.8B \citep{qwen3.5} to predict a bounding
box from an image and a referring expression. It retains the pretrained vision encoder,
visual merger, and language backbone, and maps a final hidden state to
four continuous coordinates through a lightweight regression head.
The model is trained with geometric supervision, first adapting the
new head and then jointly updating the head and part of the backbone.
Figure~\ref{fig:framework} summarizes the prediction pipeline, geometric
supervision, and performance of \method{}.

\subsection{Task Formulation}
\label{sec:method_input}
Given an image $I$ of width $W_I$ and height $H_I$, and an expression
$q$, the task is to localize the referred object with one axis-aligned
bounding box. Let $\mathbf r=(x_1,y_1,x_2,y_2)^{\mathsf T}$ denote
the annotated top-left and bottom-right corners, normalized by the
original image width and height. We represent the target as
$\mathbf b=(c_x,c_y,w,h)^{\mathsf T}\in[0,1]^4$, where
\begin{equation}
\begin{aligned}
 c_x&=\frac{x_1+x_2}{2}, & c_y&=\frac{y_1+y_2}{2},\\
 w&=x_2-x_1, & h&=y_2-y_1.
\end{aligned}
\label{eq:target_box}
\end{equation}
\subsection{Multimodal Representation}
\label{sec:method_representation}
The vision encoder processes the image, and the visual merger maps
its features to the language backbone's input representation. The
language backbone processes the resulting visual and textual sequence
using the pretrained architecture and attention structure. Denote the
final hidden sequence by
\begin{equation}
 \mathbf H=f_{\theta}(I,q)\in\mathbb R^{L\times d},
\label{eq:hidden_sequence}
\end{equation}
where $f_{\theta}$ denotes the multimodal backbone, $L$ is the padded
sequence length, and $d$ is the backbone's hidden dimension.
Let $m_t\in\{0,1\}$ be the original input mask, with one indicating
a valid position. We select
\begin{equation}
 j=\max\{t:m_t=1\},\qquad \mathbf z=\mathbf H_{j,:}^{\mathsf T}.
\label{eq:last_valid_state}
\end{equation}
This position follows the other valid visual and textual input positions,
allowing its representation to incorporate their information through the
backbone. Selecting it from the mask gives the same
definition for left and right padding. We use this existing input
position without adding a task-specific regression token.

\subsection{Bounding Box Prediction}
\label{sec:method_prediction}
The regression head $g_{\phi}$ maps $\mathbf z$ to the normalized box
through layer normalization, a hidden projection with GELU, and a
four-dimensional output projection with an elementwise sigmoid:
\begin{equation}
\begin{aligned}
 \mathbf u&=\operatorname{GELU}
    \bigl(\mathbf W_1\operatorname{LN}(\mathbf z)+\mathbf a_1\bigr),\\
 \hat{\mathbf b}&=\sigma(\mathbf W_2\mathbf u+\mathbf a_2).
\end{aligned}
\label{eq:box_head}
\end{equation}
Here $\mathbf W_1\in\mathbb R^{d_h\times d}$ and
$\mathbf W_2\in\mathbb R^{4\times d_h}$, where $d_h$ is the head's
hidden dimension; $\mathbf a_1$ and $\mathbf a_2$ are the corresponding
bias vectors. The parameters $\phi$ include both projections and layer
normalization. All four coordinates are predicted together from
the shared representation. Prediction bypasses the pretrained
vocabulary projection.

The sigmoid constrains the predicted center and dimensions to the
unit range. The resulting corners can extend beyond the image.
We define the conversion from center--size to
corner coordinates as
\begin{equation}
 \Gamma(\mathbf b)=
 \begin{bmatrix}
 c_x-w/2\\ c_y-h/2\\ c_x+w/2\\ c_y+h/2
 \end{bmatrix}.
\label{eq:box_corners}
\end{equation}
This conversion is shared by the geometric objective and inference;
boundary clipping is applied only when producing the final prediction.

\subsection{Training Objective}
\label{sec:method_objective}
We supervise both the box parameters and the overlap of the predicted
and annotated regions. For a minibatch of $B$ examples, the coordinate
loss is
\begin{equation}
 \mathcal L_{\mathrm{L1}}=
 \frac{1}{B}\sum_{n=1}^{B}\sum_{k=1}^{4}
 \bigl|\hat b_{n,k}-b_{n,k}\bigr|.
\label{eq:coordinate_loss}
\end{equation}
Thus, coordinate errors are summed within each box and averaged across
examples. To account for box overlap and enclosure, we also use
generalized intersection over union (GIoU)
\citep{rezatofighi2019giou}. Let $P$ and $G$ be the rectangular regions
specified by $\Gamma(\hat{\mathbf b})$ and $\Gamma(\mathbf b)$.
Write $a_{\cap}$ and $a_{\cup}$ for their intersection and union areas,
and $a_C$ for the area of the smallest axis-aligned rectangle enclosing
both. The numerically stabilized GIoU is
\begin{equation}
 \operatorname{GIoU}(P,G)=
 \frac{a_{\cap}}{\max(a_{\cup},\epsilon)}
 -\frac{a_C-a_{\cup}}{\max(a_C,\epsilon)},
\label{eq:giou}
\end{equation}
where $\epsilon>0$ ensures numerical stability. Each prediction is
compared with its corresponding annotated box. The overlap loss and
total objective are
\begin{equation}
\begin{aligned}
 \mathcal L_{\mathrm{GIoU}}
   &=\frac{1}{B}\sum_{n=1}^{B}
       \bigl[1-\operatorname{GIoU}(P_n,G_n)\bigr],\\
 \mathcal L&=\lambda_{\mathrm{L1}}\mathcal L_{\mathrm{L1}}
       +\lambda_{\mathrm{GIoU}}\mathcal L_{\mathrm{GIoU}}.
\end{aligned}
\label{eq:training_loss}
\end{equation}
The fixed positive coefficients $\lambda_{\mathrm{L1}}$ and
$\lambda_{\mathrm{GIoU}}$ balance the two loss terms.
The first term directly supervises normalized centers and dimensions;
the second couples these quantities through the resulting box geometry.
We compute GIoU on the un-clipped corner boxes, so out-of-image extent
remains part of the training penalty. This geometric objective is the
sole training loss.

\subsection{Adaptation and Inference}
\label{sec:method_adaptation}
The backbone is initialized from Qwen3.5-0.8B, and the regression head
is newly initialized. Training proceeds in two stages with the same
objective in Eq.~\ref{eq:training_loss}. In the first stage, we freeze
the entire backbone and optimize only $\phi$. This fits the added head
to the pretrained representations before allowing the localization loss
to update the backbone. In the second stage, we retain the adapted
head and jointly optimize it with the language backbone, including its
input embeddings, and the visual merger. The remaining vision encoder
parameters stay frozen.

At inference, we construct the same image--expression input, select the
state in Eq.~\ref{eq:last_valid_state}, and predict
$\hat{\mathbf b}$ using Eq.~\ref{eq:box_head}. We convert it to
corner coordinates and clip each coordinate to $[0,1]$:
\begin{equation}
\begin{aligned}
 \hat{\mathbf r}&=\operatorname{clip}
       \bigl(\Gamma(\hat{\mathbf b}),0,1\bigr),\\
 \hat{\mathbf r}_{\mathrm{pixel}}&=
       (W_I,H_I,W_I,H_I)^{\mathsf T}\odot\hat{\mathbf r},
\end{aligned}
\label{eq:inference_box}
\end{equation}
where $\odot$ denotes elementwise multiplication. The resulting box
is expressed in the original image's pixel coordinates. Prediction
requires one evaluation of the complete multimodal backbone followed
by the regression head and this coordinate conversion. Visual encoding
and image--text input processing are retained; the four box values
are returned without sequential coordinate-token generation or text
parsing.

\section{Experiments}
\label{sec:setup}
\label{sec:results}

\subsection{Experimental Setup}
\paragraph{Benchmarks and metric.}
We use the official training and test splits of RefCOCO, RefCOCO+, and
RefCOCOg \citep{yu2016modelingcontext,mao2016generation}.
The primary metric is \acc{}, the percentage of predictions with IoU
at least 0.5. Reported gains are absolute accuracy differences.

\paragraph{Main comparison.}
\textbf{Base} denotes Qwen3.5-0.8B
\citep{qwen3.5}, without our task-specific fine-tuning.
\textbf{\method{}} is our model trained on the combined training
partitions of the three benchmarks. We compare it with Base to assess
complete task adaptation, and with the released \textbf{LocateAnything}
model to assess grounding performance against a specialized model
\citep{wang2026locateanything}. We additionally include the released
\textbf{Hi-Token} model in the inference-cost comparison
\citep{zhu2026hitoken}. The output-formulation ablations
compare fine-tuned models. The component ablations and multi-seed
output comparison use a shared training subset.

\subsection{Main Results}
\label{sec:main_results}

\begin{table}[!t]
\centering\small
\begingroup%
\setlength{\tabcolsep}{2pt}%
\renewcommand{\arraystretch}{1.10}%
\begin{tabular*}{\linewidth}{@{\extracolsep{\fill}}lrrrrr@{}}
\toprule
\multirow{2}{*}{Model} & \multicolumn{2}{c}{RefCOCO}
& \multicolumn{2}{c}{RefCOCO+} & RefCOCOg \\
\cmidrule(lr){2-3}\cmidrule(lr){4-5}\cmidrule(l){6-6}
& testA & testB & testA & testB & test \\
\midrule
Base & 84.27 & 74.72 & 76.53 & 62.57 & 77.96 \\
\rowcolor[HTML]{EEF4F8}
\textbf{\method{}} & \textbf{93.90} & 88.36 & \textbf{90.69} & \textbf{80.08} & 87.72 \\
$\Delta$ vs. Base
& \textcolor[HTML]{D55E00}{+9.63}
& \textcolor[HTML]{D55E00}{+13.64}
& \textcolor[HTML]{D55E00}{+14.16}
& \textcolor[HTML]{D55E00}{+17.51}
& \textcolor[HTML]{D55E00}{+9.76} \\
\midrule
NExT-Chat & 89.66 & 77.04 & 83.76 & 66.19 & 79.28 \\
LocateAnything & 93.23 & \textbf{89.26} & 88.00 & 79.57 & \textbf{88.54} \\
\bottomrule
\end{tabular*}%
\endgroup%
\caption{Grounding accuracy across five test splits in \acc{} (\%).
The shaded row highlights \method{}; $\Delta$ reports its improvement
over Base in accuracy points. Bold denotes the highest model accuracy
in each column.}
\label{tab:main_accuracy}
\end{table}

\method{} improves grounding accuracy over Base on every test split,
with gains ranging from 9.63 to 17.51 accuracy points
(Table~\ref{tab:main_accuracy}). The improvements are larger on the
initially weaker testB splits of RefCOCO and RefCOCO+.
The strongest gains occur on RefCOCO+, where annotation restricts
location words \citep{yu2016modelingcontext}.
Improvement on both RefCOCO+ splits shows that the adaptation remains
effective when explicit location words are restricted, extending the
gains across different expression annotation settings.

The advantage also extends to NExT-Chat, which already predicts
continuous boxes through a regression head. \method{} outperforms it
on every split, with gains of 4.24 to 13.89 accuracy points and the
largest gap on RefCOCO+ testB. The comparison shows that accurate
continuous box prediction is achievable with the compact backbone
used by \method{}.

Against LocateAnything, \method{} achieves the highest \acc{} on
three of the five splits, including both RefCOCO+ splits.
LocateAnything scores higher on RefCOCO testB and RefCOCOg.
Across all test expressions, \method{} reaches 88.30\%, compared
with 88.00\% for LocateAnything and 79.57\% for NExT-Chat.
The compact model thus combines consistent improvements over Base
with accuracy comparable to LocateAnything and higher accuracy than
NExT-Chat across all five splits.

\begin{figure}[t]
\centering
\includegraphics[width=\columnwidth]{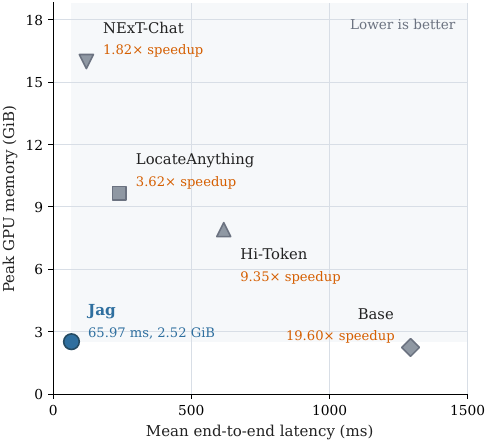}
\caption{\method{} achieves the lowest end-to-end latency among the
compared models and uses less GPU memory than NExT-Chat,
LocateAnything, and Hi-Token. Annotations report its speedup over each reference model;
the shaded region denotes combinations with
both higher latency and higher memory use than \method{}.}
\label{fig:inference_cost}
\end{figure}

\begin{figure*}[!t]
\centering\includegraphics[width=\textwidth]{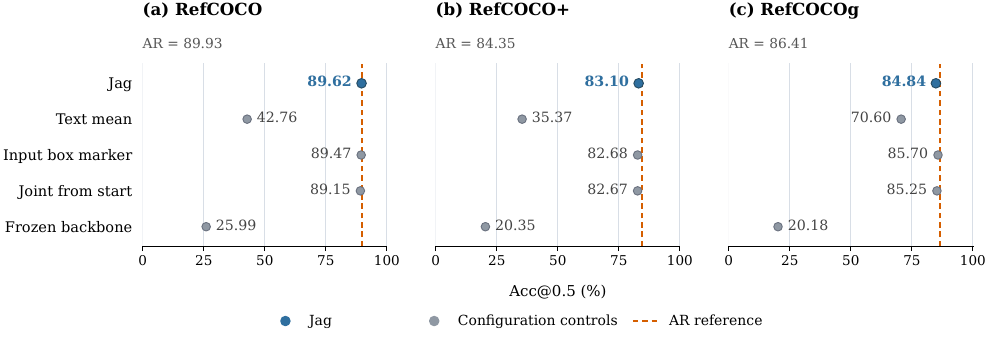}
\caption{Joint adaptation of the language backbone, visual merger, and head
outperforms head-only training. Panels compare readout and training
variants using \acc{}. Gray points are configuration
controls; orange dashed lines show the AR reference.}
\label{fig:components}
\end{figure*}

\begin{figure*}[!t]
\centering\includegraphics[width=\textwidth]{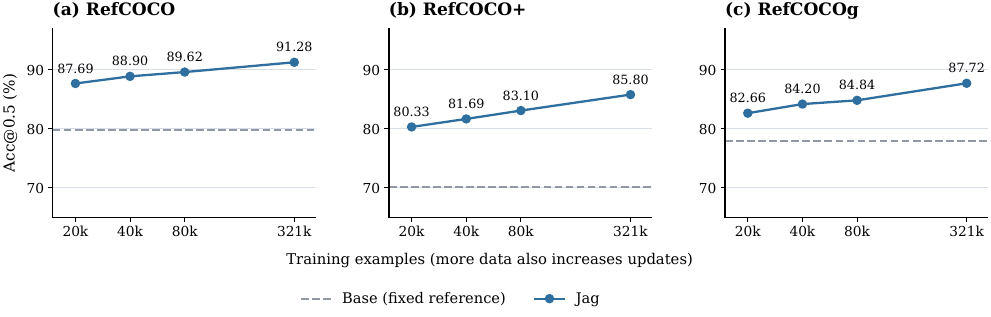}
\caption{Increasing the training budget improves \method{} on all three
datasets. Blue curves show \acc{}; gray dashed lines mark
the accuracy of Base. The example and update budgets vary together.}
\label{fig:base_budget}
\end{figure*}

\paragraph{Inference efficiency.}
\method{} achieves the lowest measured mean end-to-end latency in
single-request inference, 65.97\,ms
(Figure~\ref{fig:inference_cost}).
Relative to Base, it is 19.60$\times$ faster while improving grounding
accuracy on every test split. The model returns all four coordinates
after one multimodal forward pass, removing sequential coordinate
generation from the prediction process.

The efficiency advantage also extends to existing grounding models:
\method{} is 1.82$\times$ faster than NExT-Chat, 3.62$\times$ faster
than LocateAnything, and 9.35$\times$ faster than Hi-Token
\citep{zhu2026hitoken}.
The comparison with LocateAnything combines similar overall \acc{}
with substantially shorter prediction time. Against NExT-Chat,
\method{} provides both higher accuracy on every split and lower
latency, extending the benefit to a model that also uses continuous
box regression.

Lower latency is accompanied by a small GPU memory footprint.
\method{} uses 2.52\,GiB of peak process memory, which is 84.22\%
lower than NExT-Chat, 73.88\% lower than LocateAnything, and 68.10\%
lower than Hi-Token. Together, the accuracy and cost measurements
show that competitive grounding accuracy can be achieved with
substantially shorter prediction time and lower GPU memory use than
the compared specialized grounding models.

\subsection{Ablation Studies}
\label{sec:adaptation}

\begin{figure*}[!htbp]
\centering\includegraphics[width=\textwidth]{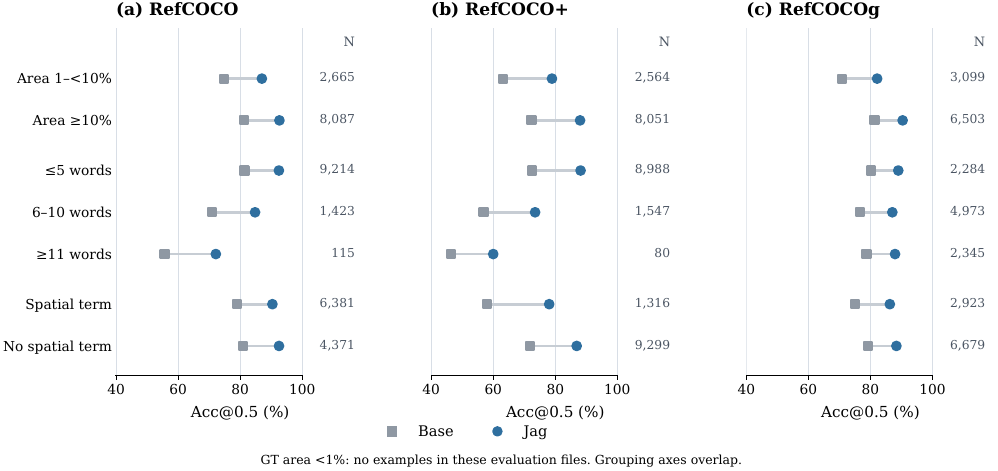}
\caption{Gains over Base occur across the observed target sizes and
expression groups. Paired points show \acc{} and counts refer to
expressions. Area is normalized GT box area; spatial tags are lexical
proxies. Intervals and empty groups are reported in the appendix.}
\label{fig:base_groups}
\end{figure*}

\begin{figure*}[t]
\centering\includegraphics[width=\textwidth]{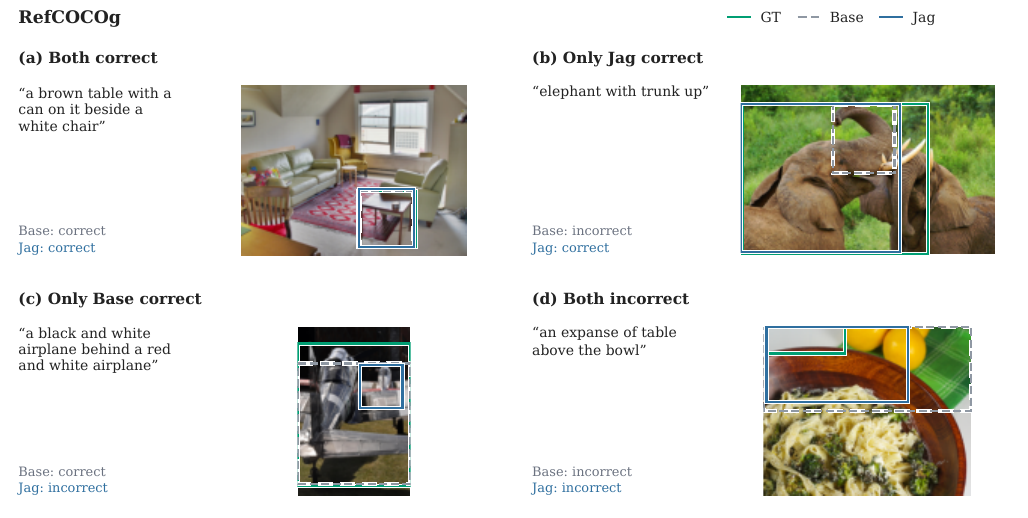}
\caption{Individual gains and failures in the Base comparison. Four RefCOCOg examples cover shared success, success only
for \method{}, success only for Base, and shared failure at IoU 0.5.
Cases are selected near each outcome's median prediction pair.}
\label{fig:base_cases}
\end{figure*}

\paragraph{Backbone adaptation.}
The component ablations show that adapting the representations is
important for accurate direct prediction (Figure~\ref{fig:components}).
Jointly updating the language backbone, visual merger, and regression
head improves \acc{} by more than 60 points on each dataset relative
to head-only training. In this training setting, fitting a continuous
box head to fixed multimodal features falls well short of adapting
those features jointly with the head.

\paragraph{Readout and training schedule.}
Averaging the referring-expression token states substantially reduces
accuracy, whereas an added input marker remains within one point of
the default on all three datasets. The last valid input state therefore
provides a competitive readout without a dedicated marker.
Immediate joint training also stays within one point of the default,
with no consistent advantage for the initial head-only stage.
The adaptation gain depends more strongly on updating the backbone
than on either of these readout or scheduling choices.

\label{sec:ablations}
\begin{table}[!t]
\centering\small
\begingroup%
\setlength{\tabcolsep}{4pt}%
\begin{tabular*}{\linewidth}{@{\extracolsep{\fill}}lrrr@{}}
\toprule
Dataset & AR & \shortstack[r]{Parallel\\discrete} & \method{} \\
\midrule
RefCOCO & 90.02 & 73.52 & 89.57 \\
 & $\pm\,0.12$ & $\pm\,1.88$ & $\pm\,0.07$ \\[3pt]
RefCOCO+ & 84.57 & 68.13 & 83.22 \\
 & $\pm\,0.31$ & $\pm\,1.26$ & $\pm\,0.17$ \\[3pt]
RefCOCOg & 86.41 & 68.04 & 85.12 \\
 & $\pm\,0.01$ & $\pm\,1.45$ & $\pm\,0.40$ \\
\bottomrule
\end{tabular*}%
\endgroup%
\caption{Output-formulation ablations: \acc{} mean $\pm$ sample
standard deviation across multiple random seeds. Parallel discrete uses four
independent categorical heads.}
\label{tab:formulations}
\end{table}

\paragraph{Continuous and discrete prediction.}
The benefit of direct prediction also depends on how coordinates are
represented and supervised. With a shared backbone and readout,
continuous regression outperforms four independent categorical heads
on all three datasets across multiple random seeds
(Table~\ref{tab:formulations}). Both models predict the coordinates
simultaneously, so simultaneous output alone is insufficient to match
the continuous model's accuracy. This comparison favors continuous
geometric prediction over the independent categorical formulation;
it changes quantization and supervision together.

\begin{table}[!t]
\centering\small
\begingroup%
\setlength{\tabcolsep}{3pt}%
\begin{tabular*}{\linewidth}{@{\extracolsep{\fill}}lrrr@{}}
\toprule
Metric & RefCOCO & RefCOCO+ & RefCOCOg \\
\midrule
AR & 90.93 & 85.82 & 88.48 \\
\method{} & 91.28 & 85.80 & 87.72 \\
$\Delta$ & +0.34 & -0.02 & -0.76 \\
95\% CI & \shortstack[r]{[-0.25,\\+0.93]} & \shortstack[r]{[-0.73,\\+0.70]} & \shortstack[r]{[-1.41,\\-0.13]} \\
\bottomrule
\end{tabular*}%
\endgroup%

\caption{Full-data comparison with the fine-tuned autoregressive
model (AR), in \acc{} (\%). Differences and paired 95\% intervals
are \method{} minus AR.}
\label{tab:full_data}
\end{table}

\paragraph{Comparison with autoregressive prediction.}
With the full training set, \method{} remains within one accuracy
point of the fine-tuned autoregressive model (AR) on every dataset
(Table~\ref{tab:full_data}). The paired difference intervals include
zero on RefCOCO and RefCOCO+, while RefCOCOg favors AR.
The multi-seed comparison likewise gives AR a higher mean accuracy
(Table~\ref{tab:formulations}). At IoU 0.75 and 0.9, AR is more accurate on all three datasets,
indicating a larger difference in
precise box placement than in \acc{}.
A separate within-backbone timing comparison gives \method{} an
8.90$\times$ mean end-to-end speedup over AR for single-request
inference. Direct prediction thus retains numerically close accuracy
at the standard threshold while substantially reducing prediction time.

\subsection{Effect of Training Data Budget}
\label{sec:scaling}

The adaptation continues to benefit from larger training budgets
(Figure~\ref{fig:base_budget}). The smallest subset already exceeds
Base on all three benchmarks, and accuracy keeps improving through
the full training set. RefCOCO+ shows the largest increase, from
80.33\% to 85.80\%. These curves describe the combined benefit of
more training examples and more optimizer updates.

\subsection{Diagnostic Analysis: Groups and Cases}
\label{sec:diagnostics}

\paragraph{Accuracy across sample groups.}
The improvements over Base extend to every nonempty group defined by
target area, expression length, and spatial keywords
(Figure~\ref{fig:base_groups}). For example, accuracy on RefCOCOg
medium-area targets increases from 70.96\% to 82.16\%, with a larger
gain than on large targets. RefCOCO+ expressions containing spatial
keywords also show larger gains than those without them.
Expression length yields no consistent gradient in the gains across
datasets. 
\paragraph{Target selection and box extent.}
The qualitative cases illustrate two kinds of prediction difference
(Figure~\ref{fig:base_cases}). In the elephant example, Base encloses
the raised trunk, whereas \method{} covers the referred animal.
In the airplane example, \method{} selects a different airplane from
the annotated target, while Base localizes it correctly. Both models
overestimate the requested table region above the bowl.
These cases distinguish selecting the wrong instance from predicting
an incorrect extent for the requested object; either can produce a
failure at IoU 0.5.

\section{Conclusion}
We presented \method{}, which adapts a pretrained multimodal model
for direct bounding-box prediction. The model predicts continuous
coordinates from the last valid input-token state, returning a complete
box in one forward pass while retaining the backbone's image--language
processing. Geometric supervision and joint adaptation make this
existing representation suitable for localization. Our component studies
show that updating the language backbone and visual merger alongside
the regression head is important for accurate prediction.

Across the evaluated benchmarks, \method{} consistently improves
grounding accuracy over Base and is competitive with existing grounding
models, with substantially lower single-request latency and low GPU
memory use. Comparisons with a fine-tuned autoregressive model show
numerically close accuracy at the standard overlap criterion and a
trade-off in precise box placement at stricter criteria. Direct prediction
therefore offers a practical choice when responsiveness matters alongside
target localization. More broadly, the study shows how the geometric
structure of visual grounding can guide both representation adaptation
and the model's output, enabling efficient localization with a compact
multimodal backbone.

\section*{Limitations}
\method{} focuses on single-object localization in static images.
Extending this setting to multiple referents, video, or segmentation
calls for task-specific design and evaluation. The evaluation uses
natural-image referring-expression benchmarks; applications in
specialized domains should assess performance on representative data.

Inference efficiency depends on input resolution, expression length,
hardware, and deployment settings. The reported speedups describe the
measured end-to-end workloads. Applications with different input
distributions or serving configurations should measure accuracy and
latency under their intended operating conditions.

\section*{Ethical Considerations}
This study uses publicly available datasets and pretrained models.
Use of these resources and derived models should follow the applicable
licenses and attribution requirements. Applications involving images
of people should respect privacy. Since localization predictions can
be incorrect, downstream systems should include verification appropriate
to their intended use, particularly when predictions inform decisions
involving individuals.

\bibliography{custom}

@inproceedings{rezatofighi2019giou,
  title = {Generalized Intersection Over Union: A Metric and a Loss for Bounding Box Regression},
  author = {Rezatofighi, Hamid and Tsoi, Nathan and Gwak, JunYoung and Sadeghian, Amir and Reid, Ian and Savarese, Silvio},
  booktitle = {Proceedings of the IEEE/CVF Conference on Computer Vision and Pattern Recognition},
  year = {2019},
  month = jun,
  pages = {658--666},
  url = {https://openaccess.thecvf.com/content_CVPR_2019/html/Rezatofighi_Generalized_Intersection_Over_Union_A_Metric_and_a_Loss_for_CVPR_2019_paper.html}
}

@inproceedings{deng2021transvg,
  title = {{TransVG}: End-to-End Visual Grounding With Transformers},
  author = {Deng, Jiajun and Yang, Zhengyuan and Chen, Tianlang and Zhou, Wengang and Li, Houqiang},
  booktitle = {Proceedings of the IEEE/CVF International Conference on Computer Vision},
  year = {2021},
  pages = {1769--1779},
  url = {https://openaccess.thecvf.com/content/ICCV2021/html/Deng_TransVG_End-to-End_Visual_Grounding_With_Transformers_ICCV_2021_paper.html}
}

@inproceedings{kamath2021mdetr,
  title = {{MDETR} -- Modulated Detection for End-to-End Multi-Modal Understanding},
  author = {Kamath, Aishwarya and Singh, Mannat and LeCun, Yann and Synnaeve, Gabriel and Misra, Ishan and Carion, Nicolas},
  booktitle = {Proceedings of the IEEE/CVF International Conference on Computer Vision},
  year = {2021},
  pages = {1780--1790},
  url = {https://openaccess.thecvf.com/content/ICCV2021/html/Kamath_MDETR_-_Modulated_Detection_for_End-to-End_Multi-Modal_Understanding_ICCV_2021_paper.html}
}

@inproceedings{zhang2024nextchat,
  title = {{NExT-Chat}: An {LMM} for Chat, Detection and Segmentation},
  author = {Zhang, Ao and Yao, Yuan and Ji, Wei and Liu, Zhiyuan and Chua, Tat-Seng},
  booktitle = {Proceedings of the 41st International Conference on Machine Learning},
  series = {Proceedings of Machine Learning Research},
  volume = {235},
  pages = {60116--60133},
  year = {2024},
  publisher = {PMLR},
  url = {https://proceedings.mlr.press/v235/zhang24bu.html}
}

@misc{wang2026locateanything,
  title = {{LocateAnything}: Fast and High-Quality Vision-Language Grounding with Parallel Box Decoding},
  author = {Wang, Shihao and Liu, Shilong and Kuang, Yuanguo and Wei, Xinyu and Liu, Yangzhou and Li, Zhiqi and Man, Yunze and Chen, Guo and Tao, Andrew and Liu, Guilin and Kautz, Jan and Zhang, Lei and Yu, Zhiding},
  year = {2026},
  eprint = {2605.27365},
  archivePrefix = {arXiv},
  primaryClass = {cs.CV},
  doi = {10.48550/arXiv.2605.27365},
  url = {https://arxiv.org/abs/2605.27365},
  note = {arXiv preprint}
}

@inproceedings{carion2020detr,
  title = {End-to-End Object Detection with Transformers},
  author = {Carion, Nicolas and Massa, Francisco and Synnaeve, Gabriel and Usunier, Nicolas and Kirillov, Alexander and Zagoruyko, Sergey},
  booktitle = {Computer Vision -- ECCV 2020},
  series = {Lecture Notes in Computer Science},
  volume = {12346},
  pages = {213--229},
  year = {2020},
  publisher = {Springer},
  doi = {10.1007/978-3-030-58452-8_13},
  url = {https://link.springer.com/chapter/10.1007/978-3-030-58452-8_13}
}

@inproceedings{chen2022pix2seq,
  title = {{Pix2seq}: A Language Modeling Framework for Object Detection},
  author = {Chen, Ting and Saxena, Saurabh and Li, Lala and Fleet, David J. and Hinton, Geoffrey},
  booktitle = {International Conference on Learning Representations},
  year = {2022},
  url = {https://openreview.net/forum?id=e42KbIw6Wb}
}

@misc{almeida2026jev,
  title = {Introducing System One Models \& {Jev}},
  author = {Almeida, Diogo},
  year = {2026},
  month = sep,
  howpublished = {TypeSafe AI official blog},
  url = {https://typesafe.ai/blog/introducing-system-one-models-and-jev},
  urldate = {2026-09-28},
  note = {Published September 15, 2026; accessed September 28, 2026}
}

@misc{qwen3.5,
    title  = {{Qwen3.5}: Towards Native Multimodal Agents},
    author = {{Qwen Team}},
    month  = {February},
    year   = {2026},
    url    = {https://qwen.ai/blog?id=qwen3.5}
}

@inproceedings{yu2016modelingcontext,
  title = {Modeling Context in Referring Expressions},
  author = {Yu, Licheng and Poirson, Patrick and Yang, Shan and Berg, Alexander C. and Berg, Tamara L.},
  booktitle = {Computer Vision -- ECCV 2016},
  series = {Lecture Notes in Computer Science},
  volume = {9906},
  pages = {69--85},
  year = {2016},
  publisher = {Springer},
  doi = {10.1007/978-3-319-46475-6_5},
  url = {https://link.springer.com/chapter/10.1007/978-3-319-46475-6_5}
}

@inproceedings{mao2016generation,
  title = {Generation and Comprehension of Unambiguous Object Descriptions},
  author = {Mao, Junhua and Huang, Jonathan and Toshev, Alexander and Camburu, Oana and Yuille, Alan L. and Murphy, Kevin},
  booktitle = {Proceedings of the IEEE Conference on Computer Vision and Pattern Recognition},
  year = {2016},
  pages = {11--20},
  url = {https://openaccess.thecvf.com/content_cvpr_2016/html/Mao_Generation_and_Comprehension_CVPR_2016_paper.html}
}

@misc{zhu2026ioupd,
  title = {{IoU-PD}: {IoU}-Aware Privileged Distillation for Visual Grounding with Multimodal Large Language Models},
  author = {Zhu, Xiuyuan and Lu, Ke and Wu, Hao and Jiao, Siwen and Du, Zijin and Zhang, Dongming and Xue, Jian},
  year = {2026},
  month = jul,
  eprint = {2607.15732},
  archivePrefix = {arXiv},
  primaryClass = {cs.CV},
  doi = {10.48550/arXiv.2607.15732},
  url = {https://arxiv.org/abs/2607.15732v3},
  note = {Version 3, revised August 14, 2026. First submitted July 17, 2026}
}

@misc{zhu2026hitoken,
  title = {{Hi-Token}: Hierarchical Coordinate Tokenization for Generative Visual Grounding},
  author = {Zhu, Xiuyuan and Lu, Ke and Dong, Kun and Jiao, Siwen and Wu, Hao and Du, Zijin and Mao, Shun and Zhang, Dongming and Xue, Jian},
  year = {2026},
  month = aug,
  eprint = {2608.03471},
  archivePrefix = {arXiv},
  primaryClass = {cs.CV},
  doi = {10.48550/arXiv.2608.03471},
  url = {https://arxiv.org/abs/2608.03471v2},
  note = {Version 2, revised August 14, 2026. First submitted August 4, 2026}
}

@misc{chen2024revisiting,
  title = {Revisiting Referring Expression Comprehension Evaluation in the Era of Large Multimodal Models},
  author = {Chen, Jierun and Wei, Fangyun and Zhao, Jinjing and Song, Sizhe and Wu, Bohuai and Peng, Zhuoxuan and Chan, S.-H. Gary and Zhang, Hongyang},
  year = {2024},
  eprint = {2406.16866},
  archivePrefix = {arXiv},
  primaryClass = {cs.CV},
  doi = {10.48550/arXiv.2406.16866},
  url = {https://arxiv.org/abs/2406.16866}
}

@inproceedings{yu2018mattnet,
  title = {{MAttNet}: Modular Attention Network for Referring Expression Comprehension},
  author = {Yu, Licheng and Lin, Zhe and Shen, Xiaohui and Yang, Jimei and Lu, Xin and Bansal, Mohit and Berg, Tamara L.},
  booktitle = {Proceedings of the IEEE Conference on Computer Vision and Pattern Recognition},
  year = {2018},
  pages = {1307--1315},
  url = {https://openaccess.thecvf.com/content_cvpr_2018/html/Yu_MAttNet_Modular_Attention_CVPR_2018_paper.html}
}

@inproceedings{yang2019faoa,
  title = {A Fast and Accurate One-Stage Approach to Visual Grounding},
  author = {Yang, Zhengyuan and Gong, Boqing and Wang, Liwei and Huang, Wenbing and Yu, Dong and Luo, Jiebo},
  booktitle = {Proceedings of the IEEE/CVF International Conference on Computer Vision},
  year = {2019},
  pages = {4683--4693},
  url = {https://openaccess.thecvf.com/content_ICCV_2019/html/Yang_A_Fast_and_Accurate_One-Stage_Approach_to_Visual_Grounding_ICCV_2019_paper.html}
}

@inproceedings{yang2020resc,
  title = {Improving One-Stage Visual Grounding by Recursive Sub-query Construction},
  author = {Yang, Zhengyuan and Chen, Tianlang and Wang, Liwei and Luo, Jiebo},
  booktitle = {Computer Vision -- ECCV 2020},
  series = {Lecture Notes in Computer Science},
  volume = {12359},
  year = {2020},
  pages = {387--404},
  publisher = {Springer},
  doi = {10.1007/978-3-030-58568-6_23},
  url = {https://link.springer.com/chapter/10.1007/978-3-030-58568-6_23}
}

@inproceedings{huang2021lbyl,
  title = {Look Before You Leap: Learning Landmark Features for One-Stage Visual Grounding},
  author = {Huang, Binbin and Lian, Dongze and Luo, Weixin and Gao, Shenghua},
  booktitle = {Proceedings of the IEEE/CVF Conference on Computer Vision and Pattern Recognition},
  year = {2021},
  pages = {16888--16897},
  url = {https://openaccess.thecvf.com/content/CVPR2021/html/Huang_Look_Before_You_Leap_Learning_Landmark_Features_for_One-Stage_Visual_CVPR_2021_paper.html}
}

@article{deng2023transvgpp,
  title = {{TransVG++}: End-to-End Visual Grounding With Language Conditioned Vision Transformer},
  author = {Deng, Jiajun and Yang, Zhengyuan and Liu, Daqing and Chen, Tianlang and Zhou, Wengang and Zhang, Yanyong and Li, Houqiang and Ouyang, Wanli},
  journal = {IEEE Transactions on Pattern Analysis and Machine Intelligence},
  volume = {45},
  number = {11},
  year = {2023},
  pages = {13636--13652},
  doi = {10.1109/TPAMI.2023.3296823},
  url = {https://ieeexplore.ieee.org/document/10187690/}
}

@inproceedings{zhu2022seqtr,
  title = {{SeqTR}: A Simple Yet Universal Network for Visual Grounding},
  author = {Zhu, Chaoyang and Zhou, Yiyi and Shen, Yunhang and Luo, Gen and Pan, Xingjia and Lin, Mingbao and Chen, Chao and Cao, Liujuan and Sun, Xiaoshuai and Ji, Rongrong},
  booktitle = {Computer Vision -- ECCV 2022},
  series = {Lecture Notes in Computer Science},
  volume = {13695},
  year = {2022},
  pages = {598--615},
  publisher = {Springer},
  doi = {10.1007/978-3-031-19833-5_35},
  url = {https://link.springer.com/chapter/10.1007/978-3-031-19833-5_35}
}

@misc{chen2023shikra,
  title = {{Shikra}: Unleashing Multimodal {LLM}'s Referential Dialogue Magic},
  author = {Chen, Keqin and Zhang, Zhao and Zeng, Weili and Zhang, Richong and Zhu, Feng and Zhao, Rui},
  year = {2023},
  eprint = {2306.15195},
  archivePrefix = {arXiv},
  primaryClass = {cs.CV},
  doi = {10.48550/arXiv.2306.15195},
  url = {https://arxiv.org/abs/2306.15195},
  note = {arXiv preprint}
}

@inproceedings{peng2024kosmos2,
  title = {Grounding Multimodal Large Language Models to the World},
  author = {Peng, Zhiliang and Wang, Wenhui and Dong, Li and Hao, Yaru and Huang, Shaohan and Ma, Shuming and Ye, Qixiang and Wei, Furu},
  booktitle = {International Conference on Learning Representations},
  pages = {51575--51598},
  year = {2024},
  url = {https://proceedings.iclr.cc/paper_files/paper/2024/hash/e112a4671e8779aa9f640a0e3f81bd26-Abstract-Conference.html}
}

@inproceedings{you2024ferret,
  title = {{Ferret}: Refer and Ground Anything Anywhere at Any Granularity},
  author = {You, Haoxuan and Zhang, Haotian and Gan, Zhe and Du, Xianzhi and Zhang, Bowen and Wang, Zirui and Cao, Liangliang and Chang, Shih-Fu and Yang, Yinfei},
  booktitle = {International Conference on Learning Representations},
  pages = {57153--57180},
  year = {2024},
  url = {https://proceedings.iclr.cc/paper_files/paper/2024/hash/fd6613131889a4b656206c50a8bd7790-Abstract-Conference.html}
}

@misc{bai2023qwenvl,
  title = {{Qwen-VL}: A Versatile Vision-Language Model for Understanding, Localization, Text Reading, and Beyond},
  author = {Bai, Jinze and Bai, Shuai and Yang, Shusheng and Wang, Shijie and Tan, Sinan and Wang, Peng and Lin, Junyang and Zhou, Chang and Zhou, Jingren},
  year = {2023},
  eprint = {2308.12966},
  archivePrefix = {arXiv},
  primaryClass = {cs.CV},
  doi = {10.48550/arXiv.2308.12966},
  url = {https://arxiv.org/abs/2308.12966},
  note = {arXiv preprint}
}

@inproceedings{wang2022ofa,
  title = {{OFA}: Unifying Architectures, Tasks, and Modalities Through a Simple Sequence-to-Sequence Learning Framework},
  author = {Wang, Peng and Yang, An and Men, Rui and Lin, Junyang and Bai, Shuai and Li, Zhikang and Ma, Jianxin and Zhou, Chang and Zhou, Jingren and Yang, Hongxia},
  booktitle = {Proceedings of the 39th International Conference on Machine Learning},
  series = {Proceedings of Machine Learning Research},
  volume = {162},
  pages = {23318--23340},
  year = {2022},
  publisher = {PMLR},
  url = {https://proceedings.mlr.press/v162/wang22al.html}
}

@inproceedings{liu2025groundingdino,
  title = {{Grounding DINO}: Marrying {DINO} with Grounded Pre-training for Open-Set Object Detection},
  author = {Liu, Shilong and Zeng, Zhaoyang and Ren, Tianhe and Li, Feng and Zhang, Hao and Yang, Jie and Jiang, Qing and Li, Chunyuan and Yang, Jianwei and Su, Hang and Zhu, Jun and Zhang, Lei},
  booktitle = {Computer Vision -- ECCV 2024},
  series = {Lecture Notes in Computer Science},
  volume = {15105},
  pages = {38--55},
  year = {2025},
  publisher = {Springer Nature Switzerland},
  doi = {10.1007/978-3-031-72970-6_3},
  url = {https://link.springer.com/chapter/10.1007/978-3-031-72970-6_3}
}

@misc{liu2025visualrft,
  title = {{Visual-RFT}: Visual Reinforcement Fine-Tuning},
  author = {Liu, Ziyu and Sun, Zeyi and Zang, Yuhang and Dong, Xiaoyi and Cao, Yuhang and Duan, Haodong and Lin, Dahua and Wang, Jiaqi},
  year = {2025},
  eprint = {2503.01785},
  archivePrefix = {arXiv},
  primaryClass = {cs.CV},
  doi = {10.48550/arXiv.2503.01785},
  url = {https://arxiv.org/abs/2503.01785v1}
}

@misc{shen2025vlmr1,
  title = {{VLM-R1}: A Stable and Generalizable {R1}-style Large Vision-Language Model},
  author = {Shen, Haozhan and Liu, Peng and Li, Jingcheng and Fang, Chunxin and Ma, Yibo and Liao, Jiajia and Shen, Qiaoli and Zhang, Zilun and Zhao, Kangjia and Zhang, Qianqian and Xu, Ruochen and Zhao, Tiancheng},
  year = {2025},
  eprint = {2504.07615},
  archivePrefix = {arXiv},
  primaryClass = {cs.CV},
  doi = {10.48550/arXiv.2504.07615},
  url = {https://arxiv.org/abs/2504.07615v2}
}

@misc{jiang2025rexomni,
  title = {Detect Anything via Next Point Prediction},
  author = {Jiang, Qing and Huo, Junan and Chen, Xingyu and Xiong, Yuda and Zeng, Zhaoyang and Chen, Yihao and Ren, Tianhe and Yu, Junzhi and Zhang, Lei},
  year = {2025},
  eprint = {2510.12798},
  archivePrefix = {arXiv},
  primaryClass = {cs.CV},
  doi = {10.48550/arXiv.2510.12798},
  url = {https://arxiv.org/abs/2510.12798v1}
}

@misc{liu2023llava,
  title = {Visual Instruction Tuning},
  author = {Liu, Haotian and Li, Chunyuan and Wu, Qingyang and Lee, Yong Jae},
  year = {2023},
  eprint = {2304.08485},
  archivePrefix = {arXiv},
  doi = {10.48550/arXiv.2304.08485},
  url = {https://arxiv.org/abs/2304.08485},
  note = {arXiv preprint}
}

@inproceedings{liu2024llava15,
  title = {Improved Baselines with Visual Instruction Tuning},
  author = {Liu, Haotian and Li, Chunyuan and Li, Yuheng and Lee, Yong Jae},
  booktitle = {Proceedings of the IEEE/CVF Conference on Computer Vision and Pattern Recognition},
  pages = {26296--26306},
  year = {2024},
  url = {https://openaccess.thecvf.com/content/CVPR2024/html/Liu_Improved_Baselines_with_Visual_Instruction_Tuning_CVPR_2024_paper.html}
}

@misc{bai2025qwen25vl,
  title = {{Qwen2.5-VL} Technical Report},
  author = {Bai, Shuai and Chen, Keqin and Liu, Xuejing and Wang, Jialin and Ge, Wenbin and Song, Sibo and Dang, Kai and Wang, Peng and Wang, Shijie and Tang, Jun and Zhong, Humen and Zhu, Yuanzhi and Yang, Mingkun and Li, Zhaohai and Wan, Jianqiang and Wang, Pengfei and Ding, Wei and Fu, Zheren and Xu, Yiheng and Ye, Jiabo and Zhang, Xi and Xie, Tianbao and Cheng, Zesen and Zhang, Hang and Yang, Zhibo and Xu, Haiyang and Lin, Junyang},
  year = {2025},
  eprint = {2502.13923},
  archivePrefix = {arXiv},
  doi = {10.48550/arXiv.2502.13923},
  url = {https://arxiv.org/abs/2502.13923},
  note = {arXiv preprint}
}

@misc{bai2025qwen3vl,
  title = {{Qwen3-VL} Technical Report},
  author = {Bai, Shuai and Cai, Yuxuan and Chen, Ruizhe and Chen, Keqin and Chen, Xionghui and Cheng, Zesen and Deng, Lianghao and Ding, Wei and Gao, Chang and Ge, Chunjiang and Ge, Wenbin and Guo, Zhifang and Huang, Qidong and Huang, Jie and Huang, Fei and Hui, Binyuan and Jiang, Shutong and Li, Zhaohai and Li, Mingsheng and Li, Mei and Li, Kaixin and Lin, Zicheng and Lin, Junyang and Liu, Xuejing and Liu, Jiawei and Liu, Chenglong and Liu, Yang and Liu, Dayiheng and Liu, Shixuan and Lu, Dunjie and Luo, Ruilin and Lv, Chenxu and Men, Rui and Meng, Lingchen and Ren, Xuancheng and Ren, Xingzhang and Song, Sibo and Sun, Yuchong and Tang, Jun and Tu, Jianhong and Wan, Jianqiang and Wang, Peng and Wang, Pengfei and Wang, Qiuyue and Wang, Yuxuan and Xie, Tianbao and Xu, Yiheng and Xu, Haiyang and Xu, Jin and Yang, Zhibo and Yang, Mingkun and Yang, Jianxin and Yang, An and Yu, Bowen and Zhang, Fei and Zhang, Hang and Zhang, Xi and Zheng, Bo and Zhong, Humen and Zhou, Jingren and Zhou, Fan and Zhou, Jing and Zhu, Yuanzhi and Zhu, Ke},
  year = {2025},
  eprint = {2511.21631},
  archivePrefix = {arXiv},
  doi = {10.48550/arXiv.2511.21631},
  url = {https://arxiv.org/abs/2511.21631},
  note = {arXiv preprint}
}

@misc{zhu2025internvl3,
  title = {{InternVL3}: Exploring Advanced Training and Test-Time Recipes for Open-Source Multimodal Models},
  author = {Zhu, Jinguo and Wang, Weiyun and Chen, Zhe and Liu, Zhaoyang and Ye, Shenglong and Gu, Lixin and Tian, Hao and Duan, Yuchen and Su, Weijie and Shao, Jie and Gao, Zhangwei and Cui, Erfei and Wang, Xuehui and Cao, Yue and Liu, Yangzhou and Wei, Xingguang and Zhang, Hongjie and Wang, Haomin and Xu, Weiye and Li, Hao and Wang, Jiahao and Deng, Nianchen and Li, Songze and He, Yinan and Jiang, Tan and Luo, Jiapeng and Wang, Yi and He, Conghui and Shi, Botian and Zhang, Xingcheng and Shao, Wenqi and He, Junjun and Xiong, Yingtong and Qu, Wenwen and Sun, Peng and Jiao, Penglong and Lv, Han and Wu, Lijun and Zhang, Kaipeng and Deng, Huipeng and Ge, Jiaye and Chen, Kai and Wang, Limin and Dou, Min and Lu, Lewei and Zhu, Xizhou and Lu, Tong and Lin, Dahua and Qiao, Yu and Dai, Jifeng and Wang, Wenhai},
  year = {2025},
  eprint = {2504.10479},
  archivePrefix = {arXiv},
  doi = {10.48550/arXiv.2504.10479},
  url = {https://arxiv.org/abs/2504.10479},
  note = {arXiv preprint}
}

@misc{wu2024deepseekvl2,
  title = {{DeepSeek-VL2}: Mixture-of-Experts Vision-Language Models for Advanced Multimodal Understanding},
  author = {Wu, Zhiyu and Chen, Xiaokang and Pan, Zizheng and Liu, Xingchao and Liu, Wen and Dai, Damai and Gao, Huazuo and Ma, Yiyang and Wu, Chengyue and Wang, Bingxuan and Xie, Zhenda and Wu, Yu and Hu, Kai and Wang, Jiawei and Sun, Yaofeng and Li, Yukun and Piao, Yishi and Guan, Kang and Liu, Aixin and Xie, Xin and You, Yuxiang and Dong, Kai and Yu, Xingkai and Zhang, Haowei and Zhao, Liang and Wang, Yisong and Ruan, Chong},
  year = {2024},
  eprint = {2412.10302},
  archivePrefix = {arXiv},
  doi = {10.48550/arXiv.2412.10302},
  url = {https://arxiv.org/abs/2412.10302},
  note = {arXiv preprint}
}

@article{zhang2026deepgrounder,
  title = {{DeepGrounder}: Generalized Reasoning for {MLLM}-based Remote Sensing Visual Grounding},
  author = {Zhang, Jiafan and Chen, Keyan and Liu, Chenyang and Lin, Baihong and Zou, Zhengxia and Shi, Zhenwei},
  journal = {IEEE Transactions on Geoscience and Remote Sensing},
  year = {2026},
  doi = {10.1109/TGRS.2026.3697769},
  url = {https://doi.org/10.1109/TGRS.2026.3697769}
}

@misc{zhao2025swift,
      title={SWIFT:A Scalable lightWeight Infrastructure for Fine-Tuning},
      author={Yuze Zhao and Jintao Huang and Jinghan Hu and Xingjun Wang and Yunlin Mao and Daoze Zhang and Zeyinzi Jiang and Zhikai Wu and Baole Ai and Ang Wang and Wenmeng Zhou and Yingda Chen},
      year={2024},
      eprint={2408.05517},
      archivePrefix={arXiv},
      primaryClass={cs.CL},
      url={https://arxiv.org/abs/2408.05517},
}

@misc{evalscope2024,
    title={{EvalScope}: Evaluation Framework for Large Models},
    author={ModelScope Team},
    year={2024},
    url={https://github.com/modelscope/evalscope}
}

\appendix
\section{Implementation and Evaluation Details}
\label{sec:appendix_recipe}
\subsection{Data and Training Configuration}
\paragraph{Training and evaluation annotations.}
Training uses 321,327 referring-expression examples from the official
training partitions of RefCOCO, RefCOCO+, and RefCOCOg, covering 28,158
images. Evaluation uses testA and testB for RefCOCO and RefCOCO+, and
the test partition for RefCOCOg. These evaluation sets contain 10,752,
10,615, and 9,602 expressions, respectively, for a total of 30,969.
Target boxes are normalized by the original image dimensions. All
evaluation expressions contribute to the reported scores; RefCOCO and
RefCOCO+ results pool testA and testB by expression count.

\paragraph{Optimization.}
Training uses ModelScope ms-swift and scoring uses EvalScope, with
dependencies specified in Docker. Head adaptation uses learning rate
$5\times10^{-4}$, ten warmup updates, and cosine decay. Joint training
uses $10^{-5}$ for the language backbone and visual merger, and
$10^{-4}$ for the regression head, with 5\% warmup and cosine decay.
AdamW uses $(\beta_1,\beta_2)=(0.9,0.999)$, $\epsilon=10^{-8}$,
weight decay 0.01, and gradient clipping at 1.0. The vision encoder
outside the merger remains frozen. Full-data training uses an effective
batch size of 96. Evaluation uses the final trained models;
the test sets are excluded from model selection. The two stages use
separate optimizers and schedulers, retaining the adapted head when
joint training begins. Model weights are stored in FP32, and backbone
computation uses BF16 autocast. Full-data, data-budget, and model-scale
results, together with the readout and adaptation ablations, report
individual training runs. The
output-formulation comparison evaluates AR, continuous regression,
and parallel discrete prediction across multiple random seeds.

\paragraph{Head configuration.}
For the 0.8B backbone, $d=1024$ and $d_h=512$. Layer normalization,
the two affine projections, GELU, and sigmoid give 528,900 parameters,
including biases and normalization parameters. LayerNorm
uses $\epsilon=10^{-5}$. New linear weights are initialized from
$\mathcal{N}(0,0.02^2)$ with zero biases; normalization scale and bias
start at one and zero. The head and geometric losses run in FP32.
MLP logits and training corner coordinates remain unclipped;
evaluation corners are clipped to the image domain. The 0.8B readout
and adaptation ablations retain this head width and normalization.

\paragraph{Geometric loss settings.}
The coefficients in Eq.~\ref{eq:training_loss} are
$\lambda_{\mathrm{L1}}=5$ and $\lambda_{\mathrm{GIoU}}=2$.
The GIoU calculation in Eq.~\ref{eq:giou} uses $\epsilon=10^{-7}$
for numerical stability.

\paragraph{Training budgets.}
The readout and adaptation ablations, multi-seed output comparison,
and model-scale comparison use 80,000 training examples.
The total update budgets at 20k, 40k, 80k, and 321,327 examples are
518, 934, 1,768, and 6,796. Each default continuous model uses the first
100 updates for head adaptation and the remainder for joint training.
The full-data model receives 6,696 joint updates.
The training subsets are nested: 20k, 40k, 80k, and the complete set.
The budget study increases available examples and optimizer updates
together. The same data
subsets and total update counts are used by AR in the corresponding
ablation; its language backbone and visual merger are trainable throughout.

\subsection{Model Evaluation}
\paragraph{Shared grounding instruction.}
All models receive the image and the same instruction:
``Please provide the bounding box coordinate of the region this
sentence describes:'' followed by the referring expression.
Each model renders this instruction in its native conversation template.

\paragraph{Base quality protocol.}
Base is Qwen3.5-0.8B without our grounding
fine-tuning. It uses greedy generation with
cache, up to 128 new tokens, SDPA attention, and BF16 weights.
\method{} returns continuous center--size coordinates using
FP32 weights and BF16 autocast. Thinking is disabled for both.
Both models use a 262,144-pixel input limit and a 2,048-token sequence
limit. \method{} rejects inputs above the sequence limit.
The comparison evaluates complete model configurations with the shared
instruction and numerical precision specified here, using common annotations
and scoring rules described below.

\paragraph{LocateAnything quality protocol.}
We evaluate the released LocateAnything-3B model
\citep{wang2026locateanything} with BF16 weights, SDPA attention,
and its native Hybrid decoder. Coordinates within each box block are
predicted in parallel. The generation limit is 256 tokens, with
temperature 0.7, top-$p$ 0.9, and repetition penalty 1.1.
Images are resized within the
262,144-pixel budget using the model's native patch alignment.
The parser accepts one box in normalized corner coordinates on the
$[0,1000]$ scale. Sampling uses a fixed initialization for each expression.

\paragraph{NExT-Chat quality protocol.}
We evaluate the released NExT-Chat-7B-336 model
\citep{zhang2024nextchat} with BF16 weights and computation.
The expression retains its original terminal punctuation, and the
native Vicuna conversation format supplies the image markers. Images are
padded to a square and processed by CLIP at $336\times336$ pixels,
producing 576 visual tokens. The official referring-expression
evaluation appends a fixed six-token answer containing \texttt{<at>}.
The MLP head predicts the box from this token's last-layer hidden state
in one forward pass, without token generation. Predicted corners are
mapped from the padded square to the original image before scoring.
All five test splits use single-example inference.

\paragraph{Input template and valid positions.}
For \method{} and AR, the processor renders the image and shared
grounding instruction using the Qwen3.5-0.8B chat template with
\texttt{add\_generation\_prompt=True} and
\texttt{enable\_thinking=False}. The last valid position belongs to
the fixed assistant prefix, after its empty thinking block; its token
decodes to two newlines. The attention mask selects the largest valid
position under either
padding direction. Direct inference uses right padding and AR uses
left padding. Readout variants use the same conversation template.

\subsection{Scoring and Statistical Analysis}
\paragraph{Output parsing and scoring.}
Base accepts one four-number JSON box, optionally fenced, with
\texttt{bbox} or \texttt{bbox\_2d} as the coordinate key; a plain array
is also accepted. AR returns one JSON array
of four center--size coordinates on the $[0,1000]$ scale.
Base and AR outputs must match these box formats; the parsers preserve
the predicted edge ordering. For every model, valid boxes are converted to normalized
corners in the original image coordinate system and clipped to $[0,1]$.
Nonfinite coordinates, malformed outputs, and zero-area predictions
score zero. Every expression remains in the denominator, and
correctness uses IoU $\geq\tau$ at threshold $\tau$. Pooled accuracies
sum correct counts and denominators across the constituent splits.

\paragraph{Confidence intervals.}
Predictions are paired by annotation identity, expression, image path,
and target box; annotation hashes identify the records.
For each dataset, we sample image clusters with replacement, keeping
all referring expressions for each sampled image and both models'
predictions together. We compute expression-weighted accuracies and
paired differences for each draw, then use the 2.5th and 97.5th
percentiles of 2,000 draws as the reported 95\% interval.
Common image draws are used for groups and contrasts,
so a difference between group gains is evaluated directly within each
resample. These intervals concern test-image sampling for fixed trained
models. Training variation is assessed separately in the multi-seed
output-formulation comparison, which reports means and sample standard
deviations across multiple random seeds.
Subgroup contrasts are exploratory and use these paired intervals
without multiple-comparison correction.

\section{Detailed Base Comparison}
\label{sec:appendix_base}
\subsection{Accuracy and Paired Outcomes}
The gains over Base have positive paired image-cluster intervals on
all three benchmarks (Table~\ref{tab:base_pooled}).
Table~\ref{tab:base_splits} gives split accuracies and sample counts;
Table~\ref{tab:base_outcomes} decomposes the comparison into four
mutually exclusive correctness outcomes. Subtracting Base-only from
\method{}-only successes gives net increases of 1,240, 1,667, and 937
correct expressions on RefCOCO, RefCOCO+, and RefCOCOg, respectively.
These gains exceed the 53, 111, and 78 invalid outputs produced by
Base; \method{} produces none.

\begin{table}[!htbp]
\centering\small
\begingroup%
\setlength{\tabcolsep}{3pt}%
\begin{tabular*}{\linewidth}{@{\extracolsep{\fill}}lrrr@{}}
\toprule
Metric & RefCOCO & RefCOCO+ & RefCOCOg \\
\midrule
Base & 79.74 & 70.10 & 77.96 \\
\method{} & 91.28 & 85.80 & 87.72 \\
$\Delta$ & +11.53 & +15.70 & +9.76 \\
95\% CI & \shortstack[r]{[+10.62,\\+12.41]} & \shortstack[r]{[+14.67,\\+16.77]} & \shortstack[r]{[+8.86,\\+10.73]} \\
\bottomrule
\end{tabular*}%
\endgroup%
\caption{Pooled Base comparison in \acc{} (\%).
$\Delta$ is \method{} minus Base, with paired image-cluster
95\% intervals.}
\label{tab:base_pooled}
\end{table}

\begin{table}[!htbp]
\centering\small\setlength{\tabcolsep}{3pt}

\begin{tabular*}{\linewidth}{@{\extracolsep{\fill}}lrrrr@{}}
\toprule
Split & $N$ & Base & \method{} & $\Delta$ \\
\midrule
\multicolumn{5}{@{}l}{RefCOCO} \\
testA & 5,657 & 84.27 & 93.90 & +9.63 \\
testB & 5,095 & 74.72 & 88.36 & +13.64 \\
\midrule
\multicolumn{5}{@{}l}{RefCOCO+} \\
testA & 5,726 & 76.53 & 90.69 & +14.16 \\
testB & 4,889 & 62.57 & 80.08 & +17.51 \\
\midrule
\multicolumn{5}{@{}l}{RefCOCOg} \\
test & 9,602 & 77.96 & 87.72 & +9.76 \\
\bottomrule
\end{tabular*}%
\caption{Split accuracies in \acc{} (\%) and sample counts for the
Base comparison. $\Delta$ denotes the absolute accuracy gain.}
\label{tab:base_splits}
\end{table}
\begin{table}[!htbp]
\centering\small\setlength{\tabcolsep}{3pt}

\begin{tabular*}{\linewidth}{@{\extracolsep{\fill}}lrrr@{}}
\toprule
Outcome & RefCOCO & RefCOCO+ & RefCOCOg \\
\midrule
Both & 8,345 & 7,063 & 7,149 \\
\method{} only & 1,469 & 2,045 & 1,274 \\
Base only & 229 & 378 & 337 \\
Neither & 709 & 1,129 & 842 \\
\bottomrule
\end{tabular*}%
\caption{Joint outcomes for the original Base comparison at IoU 0.5.
The four outcomes are mutually exclusive and cover each benchmark.}
\label{tab:base_outcomes}
\end{table}

\subsection{Subgroup and Qualitative Analysis}
\paragraph{Group definitions.}
Expression lengths are at most five, six to ten, or at least eleven
lexical words. Normalized GT areas are below 1\%, from 1\% to below
10\%, or at least 10\%. All below-1\% groups contain zero expressions,
so the area comparison covers targets occupying at least 1\% of the image.
Word counts retain internal apostrophes and hyphens within a lexical
word. The spatial proxy matches the terms ``left'', ``right'', ``above'',
``below'', ``behind'', ``beside'', ``between'', ``under'', ``underneath'',
``over'', ``near'', ``next to'', ``in front of'', ``on top of'', ``bottom'',
``top'', ``middle'', ``center'', ``closest'', ``furthest'', and ``farthest''.
These terms include positional adjectives as well as relations.
The grouping axes overlap; within each axis, groups partition the
dataset. Table~\ref{tab:app_base_groups} gives the accuracies and paired
image-cluster intervals underlying Figure~\ref{fig:base_groups}.

\begin{table*}[t]
\centering\footnotesize
\begin{tabular*}{\linewidth}{@{\extracolsep{\fill}}llrrrrr@{}}
\toprule
Dataset & Group & $N$ & Base & \method{} & $\Delta$ & 95\% CI \\
\midrule
RefCOCO & Area $<1\%$ & 0 & NA & NA & NA & NA \\
RefCOCO & Area 1--<10\% & 2,665 & 74.90 & 87.02 & +12.12 & [+10.00, +14.20] \\
RefCOCO & Area $\geq$10\% & 8,087 & 81.34 & 92.68 & +11.34 & [+10.31, +12.41] \\
RefCOCO & $\leq$5 words & 9,214 & 81.41 & 92.51 & +11.10 & [+10.18, +12.03] \\
RefCOCO & 6--10 words & 1,423 & 70.91 & 84.82 & +13.91 & [+11.19, +16.65] \\
RefCOCO & $\geq$11 words & 115 & 55.65 & 72.17 & +16.52 & [+8.21, +25.58] \\
RefCOCO & Spatial term & 6,381 & 78.97 & 90.41 & +11.44 & [+10.25, +12.60] \\
RefCOCO & No spatial term & 4,371 & 80.87 & 92.54 & +11.67 & [+10.32, +13.02] \\
RefCOCO+ & Area $<1\%$ & 0 & NA & NA & NA & NA \\
RefCOCO+ & Area 1--<10\% & 2,564 & 63.14 & 78.94 & +15.80 & [+13.38, +18.27] \\
RefCOCO+ & Area $\geq$10\% & 8,051 & 72.31 & 87.99 & +15.68 & [+14.48, +16.86] \\
RefCOCO+ & $\leq$5 words & 8,988 & 72.59 & 88.15 & +15.57 & [+14.43, +16.74] \\
RefCOCO+ & 6--10 words & 1,547 & 56.88 & 73.50 & +16.61 & [+13.97, +19.36] \\
RefCOCO+ & $\geq$11 words & 80 & 46.25 & 60.00 & +13.75 & [+1.19, +26.32] \\
RefCOCO+ & Spatial term & 1,316 & 57.98 & 78.04 & +20.06 & [+17.30, +23.13] \\
RefCOCO+ & No spatial term & 9,299 & 71.81 & 86.90 & +15.09 & [+13.97, +16.23] \\
RefCOCOg & Area $<1\%$ & 0 & NA & NA & NA & NA \\
RefCOCOg & Area 1--<10\% & 3,099 & 70.96 & 82.16 & +11.20 & [+9.45, +13.00] \\
RefCOCOg & Area $\geq$10\% & 6,503 & 81.30 & 90.37 & +9.07 & [+8.06, +10.17] \\
RefCOCOg & $\leq$5 words & 2,284 & 80.30 & 88.97 & +8.67 & [+7.11, +10.38] \\
RefCOCOg & 6--10 words & 4,973 & 76.53 & 87.05 & +10.52 & [+9.33, +11.78] \\
RefCOCOg & $\geq$11 words & 2,345 & 78.72 & 87.93 & +9.21 & [+7.52, +10.90] \\
RefCOCOg & Spatial term & 2,923 & 74.96 & 86.25 & +11.29 & [+9.68, +12.92] \\
RefCOCOg & No spatial term & 6,679 & 79.28 & 88.37 & +9.09 & [+8.09, +10.15] \\
\bottomrule
\end{tabular*}%
\caption{Base-comparison groups: \acc{} (\%), absolute gain over Base,
and paired 95\% intervals. Empty groups are retained as NA.}
\label{tab:app_base_groups}
\end{table*}

\paragraph{Group contrasts.}
On RefCOCOg, the medium-minus-large difference in gains over Base is
+2.12 points [+0.05, +4.26]. On RefCOCO+, expressions with spatial
keywords gain 20.06 points, compared with 15.09 points for expressions
without them, a contrast of +4.97 points [+1.95, +8.10]. The
long-minus-short gain intervals include zero on all three datasets.
These contrasts characterize variation within the positive gains
observed across all nonempty groups.

\paragraph{Case selection.}
Figure~\ref{fig:base_cases} presents one RefCOCOg example from each
correctness outcome. Each is the nearest case to the component-wise
median pair (Base IoU, \method{} IoU) under squared Euclidean distance,
with deterministic tie-breaking and distinct images.

\section{Supplemental Ablation Results}
\label{sec:appendix_results}
\subsection{Training, Readout, and Backbone Controls}
\paragraph{Training and output controls.}
AR fine-tunes Qwen3.5-0.8B to generate center--size coordinates as
JSON integers in $[0,1000]$ using answer-only cross entropy.
The parallel discrete control uses four independent 1,001-way
categorical heads with the same backbone and readout as continuous
regression. AR trains the original vocabulary head with the language
learning rate.
\method{} and AR use FP32 master weights with BF16 autocast.
Within each data budget, total optimizer updates, training
examples, effective batch size, and input limits are matched. Backbone
update counts, trainable parameter counts, and FLOPs differ.
At full data, AR receives 6,796 joint updates, whereas \method{} uses
100 head-only updates followed by 6,696 joint updates.

\paragraph{Readout variants.}
Table~\ref{tab:readout} gives the values underlying
Figure~\ref{fig:components}. Text mean pooling averages only positions
corresponding to the referring expression, excluding image and template
tokens. The input-marker variant appends the pretrained
\texttt{<|box\_start|>} token and reads its hidden state. The marker
is supplied as input using its existing vocabulary entry.
Immediate joint training updates the language backbone, merger, and
regression head from the first step. The frozen-backbone control
updates only the head for the full budget.

\begin{table}[!htbp]
\centering\small\setlength{\tabcolsep}{3pt}
\begin{tabular*}{\linewidth}{@{\extracolsep{\fill}}lrrr@{}}
\toprule
Configuration & RefCOCO & RefCOCO+ & RefCOCOg \\
\midrule
\method{} & 89.62 & 83.10 & 84.84 \\
Text mean & 42.76 & 35.37 & 70.60 \\
Input box marker & 89.47 & 82.68 & 85.70 \\
Joint from start & 89.15 & 82.67 & 85.25 \\
Frozen backbone & 25.99 & 20.35 & 20.18 \\
\bottomrule
\end{tabular*}%

\caption{Readout and adaptation accuracies in \acc{} (\%), underlying
Figure~\ref{fig:components}. Training uses the component-ablation budget.}
\label{tab:readout}
\end{table}

\paragraph{Backbone size.}
Table~\ref{tab:backbone_scale} compares Qwen3.5-2B models under the
same component-ablation budget. Both output formulations improve over
their 0.8B counterparts, with AR numerically higher on all three datasets.

\begin{table}[!htbp]
\centering\small\setlength{\tabcolsep}{3pt}

\begin{tabular*}{\linewidth}{@{\extracolsep{\fill}}lrr@{}}
\toprule
Dataset & AR & \method{} \\
\midrule
RefCOCO & 91.69 & 91.31 \\
RefCOCO+ & 87.33 & 85.77 \\
RefCOCOg & 89.12 & 88.28 \\
\bottomrule
\end{tabular*}%

\caption{Model-scale ablation in \acc{} (\%) using Qwen3.5-2B and
the component-ablation budget. The comparison varies model size within
the same backbone family.}
\label{tab:backbone_scale}
\end{table}

\subsection{Localization Across IoU Thresholds}
\paragraph{Localization thresholds.}
\label{sec:appendix_thresholds}
Table~\ref{tab:multithreshold} scores the same predictions on all
30,969 test expressions at multiple IoU thresholds. Following
\citet{chen2024revisiting}, mAcc averages ten accuracies at
$\tau=0.50,0.55,\ldots,0.95$, using the common scoring rules.
\method{} improves over Base at IoU 0.75 and 0.9
on all three datasets, with mAcc gains of 10.17, 14.02, and 9.60 points.
At IoU 0.95 on RefCOCO, \method{} scores 20.89\% and Base scores
22.41\%. AR scores higher than \method{} at IoU 0.75 and 0.9,
with mAcc differences of 4.24, 4.39, and 6.28 points. The comparison
distinguishes accuracy at the standard overlap criterion from precise
box placement.

\begin{table}[!htbp]
\centering\small\setlength{\tabcolsep}{3pt}

\begin{tabular*}{\linewidth}{@{\extracolsep{\fill}}lrrrr@{}}
\toprule
Model & @.5 & @.75 & @.9 & mAcc \\
\midrule
\multicolumn{5}{@{}l}{RefCOCO ($N=10,752$)}  \\
Base & 79.74 & 66.29 & 43.64 & 62.63 \\
\method{} & 91.28 & 79.06 & 50.36 & 72.80 \\
AR & 90.93 & 80.49 & 61.03 & 77.04 \\
\midrule
\multicolumn{5}{@{}l}{RefCOCO+ ($N=10,615$)}  \\
Base & 70.10 & 58.06 & 38.51 & 54.86 \\
\method{} & 85.80 & 74.88 & 48.11 & 68.88 \\
AR & 85.82 & 76.67 & 58.35 & 73.27 \\
\midrule
\multicolumn{5}{@{}l}{RefCOCOg ($N=9,602$)}  \\
Base & 77.96 & 64.31 & 42.74 & 61.03 \\
\method{} & 87.72 & 76.09 & 50.34 & 70.62 \\
AR & 88.48 & 79.29 & 64.03 & 76.91 \\
\bottomrule
\end{tabular*}%
\caption{Supplemental localization accuracy (\%). Columns @.5, @.75,
and @.9 denote IoU thresholds. mAcc averages ten thresholds from 0.50
to 0.95 at intervals of 0.05. Each dataset pools all its test expressions;
Base denotes Qwen3.5-0.8B; AR denotes its fine-tuned counterpart.}
\label{tab:multithreshold}
\end{table}

\paragraph{Expression-group comparison.}
\label{sec:appendix_diagnostics}
The same group definitions and paired image-cluster resampling give
89.98\% for AR and 87.93\% for \method{} on RefCOCOg expressions
of at least eleven words. The difference (\method{} minus AR) is
-2.05 points, with a paired 95\% interval of [-3.16, -0.87]. The
long-minus-short contrast in these model differences is -1.70 points
[-3.38, -0.08]. AR also
scores higher in the medium-area and spatial-term groups, while
their contrasts with large-area and untagged groups include zero.

\section{Inference Measurements}
\label{sec:appendix_inference}
\subsection{Model Comparison}
The five models in Figure~\ref{fig:inference_cost} process the same
evaluation inputs after separate warmup inputs, running sequentially on
one NVIDIA A100-PCIE-40GB GPU. All use BF16 backbone weights and
computation; the \method{} head retains FP32 weights and computation.
Base, \method{}, LocateAnything, and Hi-Token use SDPA and a
262,144-pixel image budget. NExT-Chat uses native $336\times336$ CLIP
preprocessing and eager attention. Unused loading allocations are
released before warmup.

CUDA-synchronized timers cover image opening, preprocessing,
host--device transfers, model execution, output decoding, and parsing,
excluding loading and warmup. Throughput divides the prediction count
by cumulative prediction time; P95 uses rank $\lceil0.95n\rceil$.
GPU memory is peak NVML process occupancy sampled every 20\,ms during
prediction. Measurements use PyTorch 2.10.0 and CUDA 12.8.

LocateAnything uses the Hybrid settings above. Hi-Token uses
Qwen2.5-VL-3B-Instruct with the shared grounding instruction,
greedy decoding, a 128-token limit, and KV
caching. The release combines coordinate supervised fine-tuning and
Hi-GAR (Hi-R1 in the original paper) \citep{zhu2026hitoken}. Its parser
retains special tokens, reconstructs four coordinates from twelve
axis-specific digits, and normalizes by 999. NExT-Chat uses its accuracy
evaluation procedure. Every input
contributes to runtime measurements, including malformed outputs.

\begin{table}[!htbp]
\centering\small
\begingroup%
\setlength{\tabcolsep}{3pt}%
\begin{tabular*}{\linewidth}{@{\extracolsep{\fill}}lrrrr@{}}
\toprule
Model & \shortstack{Mean\\(ms)} & \shortstack{P95\\(ms)}
& \shortstack{Rate\\(samples/s)} & \shortstack{Memory\\(GiB)} \\
\midrule
\method{} & 65.97 & 68.82 & 15.16 & 2.523 \\
Base & 1293.01 & 1496.69 & 0.77 & 2.246 \\
NExT-Chat & 120.01 & 141.00 & 8.33 & 15.986 \\
LocateAnything & 239.05 & 384.76 & 4.18 & 9.660 \\
Hi-Token & 616.90 & 635.55 & 1.62 & 7.910 \\
\bottomrule
\end{tabular*}%
\endgroup%
\caption{Single-request model-comparison measurements supporting
Figure~\ref{fig:inference_cost}. Latency covers complete end-to-end
prediction; rate denotes throughput and memory denotes sampled peak
GPU process occupancy.}
\label{tab:app_inference_cost}
\end{table}

\subsection{Direct and Autoregressive Prediction}
\label{sec:appendix_runtime}
Table~\ref{tab:runtime} reports the direct-versus-AR timing comparison.
\method{} and AR run on separate physical
A100-PCIE-40GB GPUs. Both process the first 256 RefCOCO testA
expressions, corresponding to 38 images, at batch sizes one and eight.
B1 warmup uses ten examples and B8 warmup uses three batches; both
reuse the measured workload. Timers include image opening,
preprocessing, host--device transfers, model execution, and output
processing, with CUDA synchronization. Model loading and warmup are
excluded. Latency percentiles use per-expression B1 times;
B8 throughput uses batched execution.
Both models use SDPA, FP32 parameters, BF16 autocast, a 262,144-pixel
input limit, and at most 2,048 input tokens. AR uses the native Qwen
generator with greedy decoding, one beam, and KV caching enabled,
stopping at EOS or 64 new tokens. The direct head runs in FP32.
Image bytes are read in a preliminary pass, and warmup revisits the
measured workload.

\begin{table}[!htbp]
\centering\small\setlength{\tabcolsep}{3pt}

\begin{tabular*}{\linewidth}{@{\extracolsep{\fill}}lrr@{}}
\toprule
Metric & AR & \method{} \\
\midrule
B1 mean (ms) & 605.80 & 68.05 \\
B1 P50 (ms) & 604.26 & 67.37 \\
B1 P95 (ms) & 623.98 & 74.28 \\
B1 (expressions/s) & 1.651 & 14.695 \\
B8 (expressions/s) & 9.954 & 35.244 \\
\bottomrule
\end{tabular*}%
\caption{Output-formulation runtime ablation. B1 and B8 denote batch sizes one
and eight; latency is in ms and throughput in expressions per second.}
\label{tab:runtime}
\end{table}

\paragraph{Latency and throughput.}
\method{} reduces mean single-request latency by 8.90$\times$ and
P95 latency from 623.98\,ms to 74.28\,ms, including image--text
processing and output computation. At batch size eight, it retains
a 3.54$\times$ throughput advantage over AR (Table~\ref{tab:runtime}).

\section{Artifact Use and Documentation}
\label{appendix:artifacts}

\subsection{Artifact Use and Licenses}
The pretrained weights, benchmark data, and software used in this study
retain the licenses and attribution requirements of their original
releases. These terms govern their use and redistribution, including
any conditions applicable to models derived from the pretrained weights.

The model files do not include the original COCO images or
referring-expression annotations; these resources are obtained
separately from their providers. Use of the adapted model and its
supporting software must also respect the applicable upstream terms.

\subsection{Artifact Documentation}
The pretrained base model for \method{} is Qwen3.5-0.8B, while
the model-scale experiments use Qwen3.5-2B \citep{qwen3.5}.
Public NExT-Chat and LocateAnything models serve as accuracy and
inference-cost comparisons; Hi-Token is included in the inference-cost
comparison
\citep{zhang2024nextchat,wang2026locateanything,zhu2026hitoken}.

Grounding examples come from RefCOCO, RefCOCO+, and RefCOCOg
\citep{yu2016modelingcontext,mao2016generation}, which pair COCO
natural images with English referring expressions and annotated boxes.
These benchmarks describe objects in general visual scenes, with one
referent to localize per expression. The study relies on their existing
annotations and introduces no new benchmark or human annotation.

ModelScope \texttt{ms-swift} provides training support
\citep{zhao2025swift}, and EvalScope provides the evaluation framework
\citep{evalscope2024}. Docker specifies the execution environment and
its software dependencies.

\subsection{AI Assistants in Research and Writing}
AI assistants helped polish the language of the manuscript.
The authors reviewed the text and retain responsibility for the
scientific content, reported results, and interpretations. They are
accountable for the accuracy and integrity of the final paper.

\end{document}